\documentclass[runningheads,a4paper]{llncs}

\usepackage{amssymb}
\usepackage{graphicx}
\usepackage{times}
\usepackage{latexsym}
\usepackage{float}
\usepackage[normalem]{ulem}
\usepackage[T1]{fontenc}
\usepackage{graphicx}
\usepackage{url}
\usepackage{subcaption}
\urldef{\mailsa}\path||
\urldef{\mailsb}\path||
\urldef{\mailsc}\path||  

\urldef{\mailsa}\path|{souvick.ghosh}@rutgers.edu|
\urldef{\mailsb}\path|{satanu.ghosh.94}@gmail.com|
\urldef{\mailsc}\path|dipankar.dipnil2005}@gmail.com|

\begin{document}

\mainmatter  % start of an individual contribution

% first the title is needed
\title{Complexity Metric for Code-Mixed Social Media Text}

% a short form should be given in case it is too long for the running head
%\titlerunning{Lecture Notes in Computer Science: Authors' Instructions}

% the name(s) of the author(s) follow(s) next
%
% NB: Chinese authors should write their first names(s) in front of
% their surnames. This ensures that the names appear correctly in
% the running heads and the author index.
%

%\author{Anonymous Submission \\}

\author{Souvick Ghosh\textsuperscript{1} \and Satanu Ghosh\textsuperscript{2}\and Dipankar Das\textsuperscript{3} \\}
%
%\authorrunning{Lecture Notes in Computer Science: Authors' Instructions}
% (feature abused for this document to repeat the title also on left hand pages)

% the affiliations are given next; don't give your e-mail address
% unless you accept that it will be published
%\institute{ Institute\\
%\mailsa\\
%\mailsb\\
%\mailsc\\}

\institute{\textsuperscript{1}Rutgers University, New Brunswick, NJ 08901, US\\
\mailsa\\
\textsuperscript{2}MAKAUT, Kolkata 700064, WB, India\\
\mailsb\\
\textsuperscript{3}Jadavpur University, Kolkata 700032, WB, India\\
\mailsc\\}
%\url{http://www.springer.com/lncs}}

\maketitle

\begin{abstract}
An evaluation metric is an absolute necessity for measuring the performance of any system and complexity of any data. In this paper, we have discussed how to determine the level of complexity of code-mixed social media texts that are growing rapidly due to multilingual interference. In general, texts written in multiple languages are often hard to comprehend and analyze. At the same time, in order to meet the demands of analysis, it is also necessary to determine the complexity of a particular document or a text segment. Thus, in the present paper, we have discussed the existing metrics for determining the code-mixing complexity of a corpus, their advantages and shortcomings as well as proposed several improvements on the existing metrics. The new index better reflects the variety and complexity of a multilingual document. Also, the index can be applied to a sentence and seamlessly extended to a paragraph or an entire document. We have employed two existing code-mixed corpora to suit the requirements of our study.
\end{abstract}

\section{Introduction}
Social media text differs from other conventional text in many ways. It is noisy in nature and requires comprehensive processing from a multilingual point of view. In social media communication, a multilingual speaker often uses more than one language. Therefore, the communication, inherently informal in nature, presents the scientific community with a challenging yet interesting problem.

First of all, we need to understand the necessity of language switching. Is it motivational? Or is it circumstantial? Although mixing of languages was prevalent in verbal communication, it was the proliferation of social media which accelerated the use of multiple languages in a single written communication and this is motivated by both social and conversational needs ~\cite{Auer:84}.  Sometimes, the speaker is not competent in the language he is writing. The lack of vocabulary provokes him to use words from his native language as a substitute. On some other occasions, the need is purely social and is used by the writer to mark him as part of a large group. 

Automatic identification of the code switching points is important as it helps to understand the frequency of code switching or code mixing and subsequent complexity of the text. Also, it would allow us to determine the language specific models which are better suited for the analysis of such text. It is also important to understand the differences between code switching and code mixing as both the terms are used interchangeably in the literature. In our work, the term 'Code- Switching' refers to inter-sentential language shifts while the term 'Code-Mixing' refers to intra-sentential shifts of language. 

In the present work, we have used short utterances collected from Facebook pages and Twitter data for our analysis. As the dataset is purely based on Indian social media text, it is essential that we give a brief statistics about the degree of multilingualism in India. There are more than 20 officially recognized languages in India. The number of Hindi speakers range from 14.5\% to 24.5\% of total population (source:Wikipedia\footnote{https://en.wikipedia.org/}). Other languages are spoken by 10\% or less out of the total population. English and Hindi are mostly used for official communication in India. Similarly, it has been also observed that the diversity of Indian languages and the necessity for faster and efficient communication motivates the mixing of languages in Indian social media context. 
This brings us to one of the challenging problems, i.e. transliteration. Most of the time, the languages like Hindi or Bengali are not written using their native scripts - Devanagari for Hindi and Eastern Neo Brahmi script for Bengali. Instead, the users prefer using Roman script as it is more convenient with a regular keyboard. While analyzing a code-mixed transliterated text, it is often useful to determine the complexity of the corpus. For any task on code-mixed corpora such as language identification, part-of-speech tagging, information retrieval or question answering, it is important for the researchers to compare the difficulty of their work with regards to the level of language mixing in the text. Also, it is expected that along with the increasing complexity and more code-mixing in a text, the accuracy of text processing would decrease and the error rates would increase.

In Section 2, we have discussed the previous work done in related area. In Section 3, we provide a brief statistics of the code-mixed corpora and its preparation. The index is presented in Section 4. The working of the index is further elaborated using various cases and examples in Section 5. Section 6 contains the results and finally, Section 7 concludes the paper and discusses the scopes of future work.

\section{Related Work}

In 2001, Kilgarriff~\cite{Kilgarriff:01} discussed and pointed out that corpus linguistics do not have proper methods for comparing corpora. Most of the corpus descriptions are textual and based on the opinions of the researchers. Such impressions are highly subjective and not a proper measure of corpus similarity or complexity. In contrast, Whenever we work on a new corpus, the questions that are inevitably raised ae about the limitations and benefits of using that corpus. The size and homogeneity of the data are some of the factors which have been used intensively. However, such approaches are mainly word based and are applicable for monolingual texts. 

Measuring corpus similarity has a wide array of applications. It has theoretical as well as research applications where one can judge the complexity of the dataset before performing their technical analysis. Also, one may want to replace a dataset with another. It is beneficial only if there is some way to determine whether the two datasets are similar and comparable in terms of their complexity and usage. This would in turn help in inter-domain portability of NLP systems too.

To the best of our knowledge, Gamb{\"a}ck and Das~\cite{Das:14} proposed the first index for code-mixed social media text. Termed as Code Mixing Index (CMI), the index tries to assess the level of code-switching in an utterance. The measure aimed at comparing one code-switched corpus with another. Gamb{\"a}ck and Das~\cite{Das:14} worked on Hindi/Bengali-English Facebook data collected from various chat groups. The corpora introduced by them has 28.5\% of the messages written in at least two languages. CMI can be described as the fraction of total words that belong to languages other than the most dominant language in the text,

\begin{center}
\begin{math}
CMI = 100*[1-\frac{max{w_i}}{n-u}]
\end{math}
 , if n > u
\newline
\begin{math}
CMI = 0
\end{math}
 , if n=u
\newline
\end{center}
where n-u is the sum of N languages present in the utterance of their respective number of words and max\{$w_i$\} is the highest number of words belonging to a particular language where n is total number of tokens and u is the number of language independent tags. Gamb{\"a}ck and Das~\cite{Das:14} averaged the CMI values for all the sentences to obtain 'CMI all' and for only the code-mixed sentences to obtain 'CMI mixed', respectively.

However, CMI considers only the fraction of words in the corpus which are code-switched. We have used CMI as an initial parameter and have suggested some improvements which would take into account the number of languages and the number of code-switching points present in the corpus.

\section{Corpora Details}

The present task requires a social media corpus which has diversity in terms of languages and code-mixed content. Forum for Information Retrieval Evaluation \footnote{http://fire.irsi.res.in/fire/2015/home} (FIRE) organized a shared task on Mixed Script Information Retrieval. The data set used for training and test suited our purpose perfectly. Another shared task was organized by the Twelfth International Conference on Natural Language Processing \footnote{http://ltrc.iiit.ac.in/icon2015/} (ICON-2015). This data set was bilingual in nature and used code-mixed social media text. Therefore, We have modified these two corpora in order to accomplish our task. 

\subsection{FIRE 2015 Shared Task Corpus}
We have modified the transliterated corpus which was provided by the organizers of FIRE 2015 Shared Task on Mixed Script Information Retrieval. The dataset contains 3701 sentences and 63526 word tokens. Each word may belong to one of the nine languages present in the entire dataset. The nine languages were Bengali (Bn), English (En), Gujarati (Gu), Hindi (Hi), Kannada (Ka), Malayalam (Ml), Marathi (Mr), Tamil (Ta) and Telugu (Te). The dataset is extremely multilingual in nature. The languages present in the dataset are the most prevalent ones that we can find in Indian social media context. It must be noted that the words of a single query usually came from 1 or 2 languages and rarely from 3 different languages. This is in line with the language mixing trends that we have witnessed in social media context. The users, even if familiar with multiple languages, rarely use more than three languages while writing their posts or tweeting. As a matter of fact, most of the sentences are bilingual in nature with one of the dominant languages as either English or Hindi.  Thus, there are sentences that mix Tamil and English words, or Bengali and Hindi words, but not for example, Gujrati and Kannada words. The named entities (marked as NE), language independent words (marked as X) and mixed words containing intra-word language switches (marked as MIX), were all considered undefined and assigned with UN (universal) tag. The numbers of utterances, tokens for each of the language pairs in the training set are given in Table \ref{table:1}.

\begin{table}[!htpb]
\small
\centering
\begin{tabular}{|p{2.5cm}|p{2.5cm}|p{2.5cm}|p{2.5cm}|}
\hline 
\bf Language Tags & \bf \# Sentences & \bf \# Words  & \bf Percentage (\%) Of Corpus \\ 
\hline
English (En) & 2665 & 21996	& 34.63 \\
\hline
Bengali (Bn) & 355 & 4919 & 7.74 \\
\hline
Gujarati (Gu) & 165 & 1075 & 1.69 \\
\hline
Hindi (Hi) & 614 & 5897 & 9.28 \\
\hline
Kannada (Ka) & 373 & 2212 & 3.48 \\
\hline
Malayalam (Ml) & 151 & 1390 & 2.19 \\
\hline
Marathi (Mr) & 229 & 2414 & 3.8 \\
\hline
Tamil (Ta) & 342 & 3694 & 5.82 \\
\hline
Telugu (Te) & 603 & 7002 & 1.1 \\
\hline
Language Independent & 2582 & 12927 & 20.35 \\
\hline
\end{tabular}
\caption{\label{font-table} Statistics of FIRE 2015 Corpus.}
\label{table:1}
\end{table}

\subsection{ICON 2015 Shared Task Corpus}
Another recent shared task was conducted by Twelfth International Conference on Natural Language Processing (ICON-2015), for part-of-speech tagging of transliterated social media text. In the shared task, the code-mixed data was collected from Bengali-English Facebook chat groups. The sentences are in mixed English-Bengali and English-Hindi - and have been obtained from the "JU Confession" Facebook group, which contains posts in English-Bengali with a few Hindi words in some cases.

We have modified the ICON Shared Task Corpora for developing our index metric. The dataset contains three languages - Bengali, Hindi and English. It contains 2341 sentences and 38199 word tokens in total. The statistics for the dataset have been presented in Table \ref{table:2}.

\begin{table}[!htpb]
\small
\centering
\begin{tabular}{|p{2.5cm}|p{2.5cm}|p{2.5cm}|p{2.5cm}|}
\hline 
\bf Language Tags & \bf Number of Sentences & \bf Number of Words Present & \bf Percentage (\%) Of Corpus \\ 
\hline
English (En) & 1563 & 15435	& 40.41 \\
\hline
Bengali (Bn) & 1059 & 13002 & 34.04 \\
\hline
Hindi (Hi) & 153 & 1006 & 2.63 \\
\hline
Language Independent & 2268 & 8756 & 22.92 \\
\hline
\end{tabular}
\caption{\label{font-table} Statistics of ICON 2015 Corpus.}
\label{table:2}
\end{table}

\section{Complexity Factor (CF)}

We introduce an index, termed hereafter as Complexity Factor (CF), to measure the complexity of a multilingual corpus. This index can be applied to any sentence, paragraph or document which contains multiple languages. The index uses the concept of CMI as proposed by Gamb{\"a}ck and Das~\cite{Das:14} and makes some practical additions on it.

Complexity Factor(CF) considers three different aspects while analyzing any text - Language Factor (LF), Switching Factor (SF) and Mix Factor (MF). CF can be calculated for sentences and easily extended to paragraphs and entire documents. In the next section, we have proposed three variants of Complexity Factor. Complexity Factor 1 (henceforth mentioned as CF1) is a simple baseline which considers LF, SF and MF. Complexity Factor 2 (CF2) and Complexity Factor 3 (CF3) are the two indexes which have been carefully fine-tuned to efficiently represent the complexity of any transliterated text.

\subsection{Language Factor (LF) }

This factor represents the number of different languages present in a sentence as a fraction of the total number of words in the sentences. It is evident that if a sentence becomes more multilingual, the complexity increases manifold. For example, 
For any given sentence, Language Factor can be defined as,

\begin{center}
\begin{math}
LF = \frac{W}{N}
\end{math}
\end{center}
where W is the number of words and N is the number of distinct languages in the sentence.
\newline
Sentence 1: "\textit{Boss, \textbf{<Bn> ajkal ki korchis </Bn>}? We have been getting no news about you!}" 
(English Translation: Boss, what are you doing in these days? We have been getting no news about you! ) 
\newline
Sentence 2: \textit{"\textbf{<Bn> Kal khela dekhli? </Bn>} What a game! \uline{<Hi> Virat ne toh kamaal kar diya! </Hi>}" }
(English translation: Did you watch the match yesterday? What a game! Virat was simply superb!)

Sentence 1 contains two languages, Bengali and English, while Sentence 2 contains three languages - English, Bengali and Hindi. In both the sentences, Bengali words are boldfaced and Hindi words are underlined.
Language Factor is 6 for Sentence 1 (W=12, N=2) and 4 for Sentence 2 (W=12, N=3). It must be noted that longer the text block we are considering, it has more probability of finding multiple languages in it. This factor is inversely proportional to Complexity Factor (CF) and rewards shorter sentences with more distinct languages in it.
The LF can range from W (in case of a monolingual text) to 1 (when each word belongs to a different language).

\subsection{Switching Factor (SF)}

It is essential to consider the number of times the writer switches from one language to the other. As the number of switches increases, it becomes more complex to analyze the text for various tasks like language identification, part-of-speech tagging, question-answering, summarization, etc. 

For any given sentence, Switching Factor is defined as the ratio of number of switching points present in the sentence to the maximum number of switching points possible for that sentence. For a block of W words, the maximum number of code-switches occurs when each alternate words belong to different languages. So the maximum number of switching points for a W-word sentence is W-1. Switching Factor, denoted by SF, can be written as:

\begin{center}
\begin{math}
SF = \frac{S}{W-1}
\end{math}
 , if W > 1
\end{center}

\begin{center}
\begin{math}
SF = 0
\end{math}
 , if W = 1
\end{center}
where S is the number of code-switches and W is the number of words in the sentences or block of text.

Consider the following examples,
\newline
Sentence 1: \textit{\textbf{Ki} post \textbf{korcho}? Public forum \textbf{eta} }
(English translation: What are you posting? This is a public forum)
\newline
Sentence 2: \textit{It is painful \textbf{je khelata harlam}}
(English translation: It is painful that we lost the match)
\newline
Both the sentences contain a mix of Bengali and English words (Bengali words are boldfaced). It should be noted that while both sentences contain 3 words each in Bengali and English, the relative arrangement of the words make Sentence 1 more complex than Sentence 2. For sentence 1, SF is 0.8 (S=4, W=6) while for sentence 2, it is only 0.2 (S=1, W=6). Thus, we can observe that Switching Factor captures this complexity factor and it is directly proportional to Complexity Factor (CF). For a single word sentence, SF = 0. SF can reach the maximum value of 1 when no two consecutive words belong to the same language.

\subsection{Mix Factor (MF) }

Mix Factor, referred to as MF for the rest of the paper, is based on Code Mixing Index (CMI). It is the ratio of number of words which are not written in the dominant language of the sentence to the total number of language-dependent words present in the sentence. It can be written as:

\begin{center}
\begin{math}
MF = \frac{W' - max\{w\}}{W'} 
\end{math}
 , if W' > 0
\end{center}

\begin{center}
\begin{math}
MF = 0 
\end{math}
 , if W' = 0
\end{center}
where W' is the number of words in distinct languages, i.e., the number of words except the undefined ones, max\{w\} is the maximum number of words belonging to the most frequent language in the sentence.
\newline
Sentence 1: \textit{"Boss, \textbf{ajkal ki korchis}? We have been getting no news about you!"}
(English Translation: Boss, what are you doing these days? We have been getting no news about you! ) 
\newline
Sentence 2: \textit{"\textbf{Kal khela dekhli}? What a game! \uline{Virat ne toh kamaal kar diya!}" }
(English translation: Did you watch the match yesterday? What a game! Virat was simply superb!)
\newline
For sentence 1, MF is 0.25 (BN: 3, EN: 9) while for sentence 2, MF is 0.5 (BN: 3, EN: 3, HI: 6)

MF can range from \begin{math} 1 - \frac{1}{W} \end{math}  (when every word in the sentence belongs to a different language) to 1 (for monolingual texts).

\subsection{Complexity Factor}

Finally, we have combined all the three factors to formulate the Complexity Factor as,
\begin{center}
\begin{math}
CF=\frac{a*MF+b*SF}{f(LF)}
\end{math}
\end{center}
where a and b are the weights for Mix Factor (MF) and Switching Factor (SF), respectively. On the other hand, f is a function of Language Factor(LF) that we use as a dampening factor.
After experimentation with the weights, we finally settled a = 50 and b = 50. We calculated CF by having f(LF) = LF, by using a linear function, \begin{math} f(LF) = (\frac{0.25}{W-1})(LF-1) + 1 \end{math}  and by using a geometric function, \begin{math} f(LF) = \frac{arctan⁡(LF)}{\pi} + 0.75 \end{math}. We have calculated CF in three different ways and presented our results in Table 3 and Table 4.

\begin{center}
\begin{math}
CF1 = \frac{50*MF+50*SF}{LF}
\end{math}
\end{center}

\begin{center}
\begin{math}
CF2 = \frac{50*MF+50*SF}{(\frac{0.25}{W-1})(LF-1) + 1}
\end{math}
\end{center}

\begin{center}
\begin{math}
CF3 = \frac{50*MF+50*SF}{\frac{arctan⁡(LF)}{\pi} + 0.75}
\end{math}
\end{center}

We have considered MF and SF to be equally important while determining the code-mixing complexity of the social media text. However, the number of languages in social media texts is often limited to two or three. So, the impact of the LF on complexity has been dampened by the use of a linear function (in CF2) and a geometric function (in CF3). This ensures that the complexity of any given text is not heavily reduced by the language factor.

\begin{figure*}[!htb]
	\centering
	\begin{subfigure}[t]{0.52\linewidth}
		\centering
        \includegraphics[width=\linewidth, height = 0.3\paperheight]{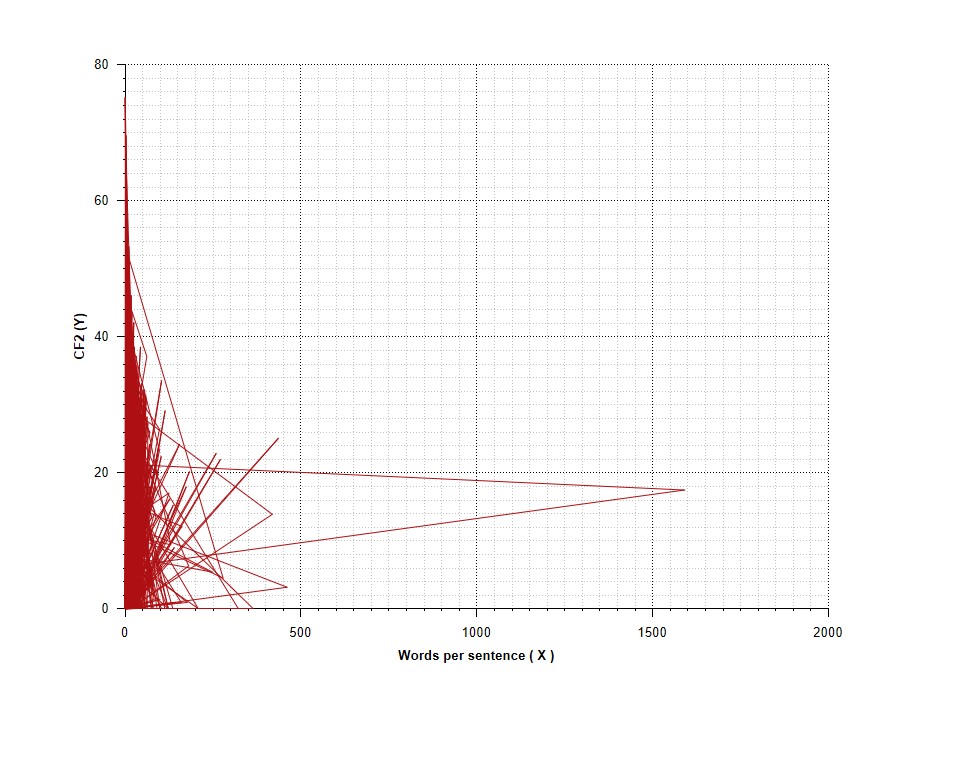}
        \caption{\label{font-figure} FIRE Corpus: Graph of Words per Sentence vs. CF2.}
        \label{fig:3}
	\end{subfigure}%
    ~
    \begin{subfigure}[t]{0.52\linewidth}
		\centering
        \includegraphics[width=\linewidth, height = 0.3\paperheight]{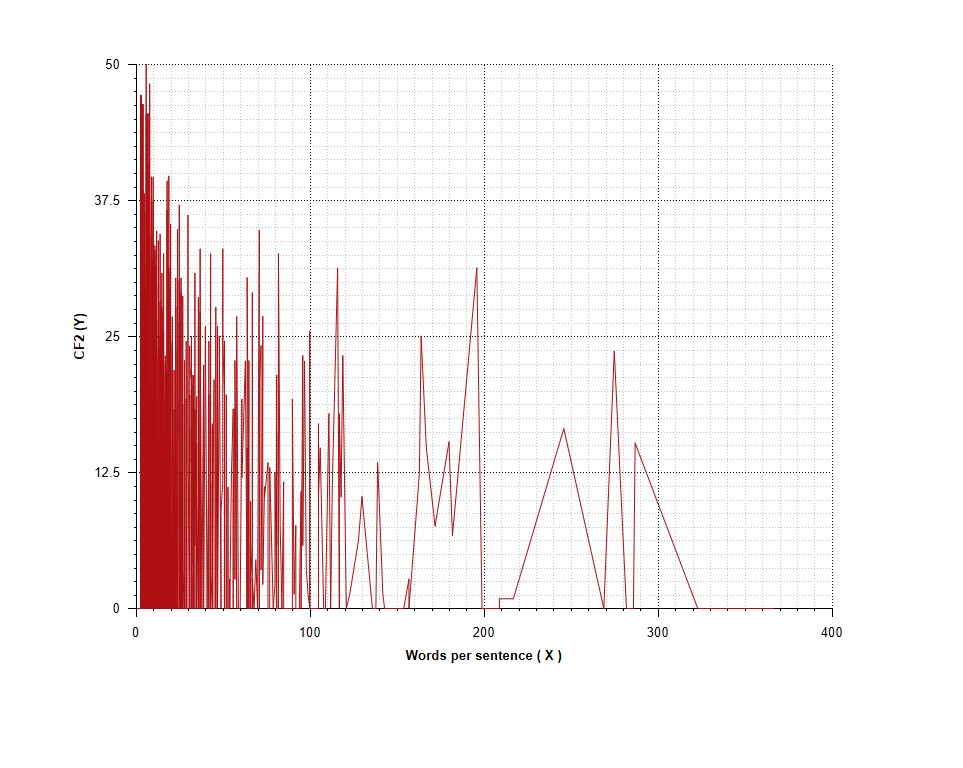}
        \caption{\label{font-figure} ICON Corpus: Graph of Words per Sentence vs. CF2.}
        \label{fig:4}
	\end{subfigure}
    \caption{Performance of Complexity Factor for both corpora}
    \label{Fig:1}
\end{figure*}

\section{Working of the Index}

We have presented a few test cases to compare the performance of our index in comparison to the existing index (CMI). The first four cases presented are from a purely mathematical perspective and serves the purpose of illustrating the mathematical precision of the index which we proposed here. For the remaining cases, we have presented examples from the dataset to elucidate our objectives. In all the examples, $w_i$ and $l_i$ represents the word and the language at position i, respectively.
\newline
\newline
\textit{Case 1: w1/l1 w2/l2 w3/l3 w4/l4 w5/l5 w6/l6 w7/l7 w8/l8 w9/l9 w10/l10} 
\newline
The sentence contains 10 words each belonging to a different language. Ideally, any index should denote the complexity of such a code-mixed sentence as 100 (in a scale of 0-100). There is no better example of a more complex sentence than this from a multilingual perspective. CMI gives a value of 90 for such a sentence. CF2 and CF3 both give the complexity as 95.
\newline
\newline
\textit{Case 2: w1/l1 w2/l1 w3/l1 w4/l1 w5/l1 w6/l1 w7/l1 w8/l1 w9/l11 w10/l1}
\newline 
The sentence contains 10 words each belonging to the same language. Here, we would expect the index to show complexity as zero. Each of the three indexes, CMI, CF2 and CF3, gives the complexity as 0. In case of CF2 and CF3, two of the three components - the language factor, switching factor and the mix factor all are zero.
\newline
\newline
\textit{Case 3: w1/l1 w2/l2 w3/l1 w4/l2 w5/l1 w6/l2 w7/l1 w8/l2 w9/l11 w10/l2}
\newline 
The sentence contains 10 words belonging to two languages. The words are arranged such that no two consecutive words belong to the same language. CMI calculates the complexity of the sentence to be 50. The complexities, as given by CF2 and CF3 are 67.5 and 63.2, respectively (LF=5, MF=0.5, SF=1).
\newline
\newline
\textit{Case 4: w1/l1 w2/l1 w3/l1 w4/l1 w5/l1 w6/l2 w7/l2 w8/l2 w9/l12 w10/l2}
\newline
The sentence contains 10 words belonging to two languages. The words are arranged such that first five words belong to the one language and the next five words belong to a second language. Once again, CMI calculates the complexity of the index as 50. CF2 and CF3 calculates the complexity to be 27.52 and 25.73, respectively (LF=5, MF=0.5, SF=0.11). It has to be mentioned that Complexity Factor correctly estimates the sentence to be less complex than the previous example (which contains more switching).
\newline \newline
The previous examples were theoretical to mainly highlight the mathematical background of the model. We have collected a few examples from our corpus to further illustrate the robustness of our index. Once again, we compared it with respect to CMI.
\newline
\newline
\textit{Case 5: Koi ni bhai , apne dbc wale hosla ni haarte ... "think to score goals instead of thinking abt goalkeepers"}
\newline
The above sentence contains 9 English and 8 Hindi words. The value of CMI is 47 while CF2 and CF3 are 23.21 and 21.31, respectively. Complexity Factor Indexes are less than the CMI because there is only one language switch in the sentence. 
\newline
\newline
\textit{Case 6: mari bike ma puncture padayu}
\newline
In the above sentence, there are 2 English and 3 Gujarati words with 4 language switches (each alternate word belongs to a different language). Complexity Factor for the sentence is 64.22 (as in CF2) and 61.75 (as in CF3) with the highest possible Code-Switch Factor (which is 1). CMI gives a value of 40 because it considers only the fraction of non-dominant languages present.
\newline
\newline
\textit{Case 7: Mi maza maharastra prem dhakvla .. tu swapnil joshi la hate karun jar saharukla support karanar asel tar saalam malakun ........... I like swapnil because he's maharastrian ... also I have never unbend opinion about you ........}
\newline
In the above sentence, there are 15 English and 15 Marathi words with 5 language switches. Although the length of the sentence is quite long, it has few language switches. CF2 and CF3 recognize that aspect and assign a complexity of 28.57 and 26.02, respectively while CMI assigns it a complexity of 50.
\newline
\newline
\textit{Case 8: Steve : 10 th anniversary celebarate pannama poiduvomo - nu .}
\newline
In this case, we have selected a smaller sentence with 3 English and 3 Tamil words. There is 1 language switch present. Once again, CMI assigns it a complexity of 50. CF2 and CF3 are 22.97 and 24.79 respectively. These values are less than that of Case 7 because of  fewer number of language switches.
\newline
\newline
\textit{Case 9: BIG B sings the eternal journey of life well .......... " tu shola ban jo khud jalke janha rashan karde ... ekla jalo re "}
\newline
This sentence contains words from 3 different languages - English (9 words), Bengali (3 words) and Hindi (9 words). The high proportion of language mixing makes CMI value 57. However, the words of all the languages occur in clusters with only 2 language switchings. Therefore, the CF2 and CF3 values are 30.09 and 26.86, respectively (as it considers the relative ordering of language words along with the presence of non-dominant language).
\newline
\newline
\textit{Case 10: Happy Rakshabandhan(Rakhi ) ...... Piyali Kar Lipika Bisht Lopamudra Sarkar Mandira Agrawal Payel Ghosh Trishona Vanhi}
\newline
This is another example which contains only two words which are language specific (1 English and 1 Bengali word). The remaining words are named entities. CMI assigns it a complexity of 50. CF2 and CF3 values are 25.45 and 23.55 respectively.
\newline
\newline
\textit{Case 11: r february te amar breakup hoy .}
\newline
This sentence contains 2 English and 4 Bengali words with 4 language switches. CMI value is 33 while CF2 and CF3 values are 45.45 and 43.1 respectively. The frequent switching of languages makes this sentence more complex than usual and Complexity Factor captures that aspect.
\newline
In the following section, we discuss the range of all the indexes in both the corpora.

\section{Results on Different Corpora}

We calculated the complexity of the FIRE and ICON corpora. The results are presented in Table \ref{table:3} and Table \ref{table:4}. The minimum and maximum values show the range of the indexes for both the corpora. It may be noted that CF2 and CF3 shows a broader range than CMI in case of FIRE corpus. The primary reason for this is the presence of more languages in the FIRE corpus. Words per sentence has been used to highlight the length of sentences present in both the corpora. The average value of the index reflects the complexity of the entire corpus.
In Figure \ref{Fig:1}, graphs have been plotted where X axis represents the length of the sentence and Y axis represents the index (CF2).

\subsection{The FIRE 2015 Shared Task Corpus}

\begin{table}[!htpb]
\small
\centering
\begin{tabular}{|p{2.8cm}|p{2.5cm}|p{2.5cm}|p{2.5cm}|}
\hline 
\bf Index & \bf Minimum Value & \bf Maximum Value & \bf Average \\ 
\hline
CMI & 0 & 50 & 11.65 \\
\hline
CF1 & 0 & 75 & 2.51 \\
\hline
CF2 & 0 & 75 & 10.54 \\
\hline
CF3 & 0 & 75 & 9.88 \\
\hline
Words/sentence & 1 & 1592 & 17.16 \\
\hline
\end{tabular}
\caption{\label{font-table} Range and Mean of Each Index and Words per Sentence (In FIRE Corpus).}
\label{table:3}
\end{table}

\subsection{The ICON 2015 Shared Task Corpus}

\begin{table}[!htpb]
\small
\centering
\begin{tabular}{|p{2.8cm}|p{2.5cm}|p{2.5cm}|p{2.5cm}|}
\hline 
\bf Index & \bf Minimum Value & \bf Maximum Value & \bf Average \\ 
\hline
CMI & 0 & 57 & 5.73 \\
\hline
CF1 & 0 & 33.5 & 1.02 \\
\hline
CF2 & 0 & 50 & 4.83 \\
\hline
CF3 & 0 & 47.38 & 4.51 \\
\hline
Words/sentence & 1 & 367 & 16.32 \\
\hline
\end{tabular}
\caption{\label{font-table} Range and Mean of Each Index and Words per Sentence (In ICON Corpus).}
\label{table:4}
\end{table}

The results suggest that the FIRE Corpus was more complex than the ICON Corpus with average value of CF2 and CF3 over the entire corpus being 10.54 and 9.88 respectively. For ICON corpus, CF2 and CF3 are 4.83 and 4.71 respectively which is considerably less than that of FIRE Corpus.

\section{Conclusion and Future Work}

In this paper, we have discussed the need and application of various indexes to represent the complexity of code-mixing in transliterated social media text. We have used various examples - both mathematical and from real-life text - to demonstrate the working of Code Mixing Index. We have highlighted few of the challenges that this index face and have proposed a new index - Complexity Factor (with two variations CF1 and CF2) - which takes into account the relative ordering of words (or the number of language switches) and the number of languages present in addition to the presence of words from non-dominant languages (as done in CMI). We have proposed three variations of Complexity Factor - CF1 serves as a baseline or raw index. CF2 and CF3 are more versatile and usable. CF2 uses linear interpolation while CF3 uses geometric functions. Both of these indexes provide a more balanced view of the complexity of the text. 
In future, the working of the index can be further explored using more data from multilingual texts (containing more than two languages). We can also find and compare the complexities of various corpora prior performing the tasks like part-of-speech tagging or sentiment analysis. The index can also be checked for mixed texts from other regions (like Spanish-English, Mandarin-English, etc.). In future, another challenging work would be to modify the index to account for complexity caused due to intra-word mixing.

\bibliography{cicling}

\begin{thebibliography}{1}
\providecommand{\url}[1]{\texttt{#1}}
\providecommand{\urlprefix}{URL }

\bibitem{Auer:84}
Auer, P.: Bilingual conversation. John Benjamins Publishing (1984)

\bibitem{Das:14}
Gamb{\"a}ck, B., Das, A.: On measuring the complexity of code-mixing. In:
  Proceedings of the 11th International Conference on Natural Language
  Processing, Goa, India. pp. 1--7 (2014)

\bibitem{Kilgarriff:01}
Kilgarriff, A.: Comparing corpora. International journal of corpus linguistics
  6(1),  97--133 (2001)

\end{thebibliography}
\bibliographystyle{cicling}

\end{document}